\documentclass{article}
\usepackage{spconf,amsmath,graphicx,subcaption}

\title{Cross domain emotion recognition using few shot knowledge transfer}
\name{Justin Olah$^{**}$, Sabyasachee Baruah$^{\dagger}$$^{\odot}$, Digbalay Bose$^{\dagger}$$^{\odot}$, Shrikanth Narayanan$^{\dagger}$}

\address{$^{**}$Stanford University\\ 
$^{\dagger}$Signal Analysis and Interpretation Laboratory, University of Southern California \\ 
\tt\small {jolah@stanford.edu,sbaruah@usc.edu,dbose@usc.edu,shri@ee.usc.edu}}

\begin{document}
\ninept
\maketitle
\def\thefootnote{$\odot$}\footnotetext{These authors contributed equally to this work}\def\thefootnote{\arabic{footnote}}
\def\thefootnote{}\footnotetext{\copyright 2022 IEEE.  Personal use of this material is permitted. Permission from IEEE must be obtained for all other uses, in any current or future media, including reprinting/republishing this material for advertising or promotional purposes, creating new collective works, for resale or redistribution to servers or lists, or reuse of any copyrighted component of this work in other works.}\def\thefootnote{\arabic{footnote}}
\begin{abstract}
Emotion recognition from text is a challenging task due to diverse emotion taxonomies, lack of reliable labeled data in different domains, and highly subjective annotation standards.
Few-shot and zero-shot techniques can generalize across unseen emotions by projecting the documents and emotion labels onto a shared embedding space. 
In this work, we explore the task of few-shot emotion recognition by transferring the knowledge gained from supervision on the GoEmotions Reddit dataset to the SemEval tweets corpus, using different emotion representation methods.
The results show that knowledge transfer using external knowledge bases and fine-tuned encoders perform comparably as supervised baselines, requiring minimal supervision from the task dataset.
\end{abstract}


\section{Introduction}
Emotion recognition in unstructured text is an important natural language understanding task, with numerous downstream applications \cite{Picard97}. 
It enables a better understanding of customers' opinions in product reviews \cite{socher-etal-2013-recursive, pontiki-etal-2014-semeval}, public sentiment in social media \cite{mohammad-etal-2018-semeval}, hate speech \cite{plazadelarco2021multitask}, political stances in news articles \cite{augenstein-etal-2016-stance}, and users' mood in chatbot interactions \cite{Lin_Xu_Winata_Siddique_Liu_Shin_Fung_2020, zhou2019design}.
Emotion recognition research has expedited several NLP benchmark datasets and tasks, particularly in the domain of social media posts and question answering forums \cite{mohammad-etal-2018-semeval, demszky2020goemotions}.
Most contemporary emotion recognition approaches involve supervised modeling using deep neural networks and transformers \cite{EmotionX, Alhuzali2021}.
However, supervised models rely on high-quality labeled data, which is difficult to acquire for subjective tasks like emotion recognition, and is noisy due to inconsistencies in the understanding of emotions between annotators of different cultural backgrounds \cite{bostan-klinger-2018-analysis}.
Moreover, emotion datasets follow different taxonomies, making it difficult to adapt existing models to newer domains and emotions.
Lately, unsupervised (few-shot and zero-shot) models have gained popularity, where a large model with billions of parameters is pretrained on extensive unlabeled text corpora with some general language modeling objective, and then optionally fine-tuned on a small amount of labeled data of the target task \cite{NIPS2013_2d6cc4b2, BERT, Brown2020}.
Zero-shot and few-shot models project the labels (emotions) and documents to a common embedding space using pretrained encoders and matrix factorization methods and compare their representations to estimate a similarity score \cite{Zad2020}.
We extend this approach to the emotion recognition task and show that models learned on one taxonomy can be easily adapted to another with minimal fine-tuning.

In this work, we explore various few-shot emotion recognition models on the GoEmotions Reddit dataset \cite{demszky2020goemotions}.
We experiment with different methods to project the emotion label to a sentence embedding.
We also show successful knowledge transfer from GoEmotions to SemEval tweets \cite{mohammad-etal-2018-semeval}, even though the two datasets follow different emotion taxonomies.
Our contributions are three-fold: 
1) we improve the supervised baseline of the GoEmotions Reddit dataset using SBERT encoder, 
2) we develop few-shot emotion recognition methods leveraging external knowledge bases like WordNet, and
3) we achieve comparable performance to the supervised baseline on SemEval tweets emotion classification by knowledge transfer from GoEmotions Reddit comments.
\vspace{-4mm}
\section{Related Work}
\textbf{Few shot and Zero shot text classification:} In the domain of zero shot text classification for social-media, Chen et.al \cite{ZSLKGtweet} rely on external knowledge-base guided projection methods for mapping posts and labels into a shared embedding space.  For unsupervised document classification, Haj-Yahia et.al \cite{haj-yahia-etal-2019-towards} perform category label enrichment through expert interventions followed by Wordnet \cite{wordnet} based definitions to compute similarities between documents and labels. We draw inspiration from these above-mentioned methods in our definition-based projection method, where we leverage emotion label definitions from Wordnet \cite{wordnet} and map it into a similar sentence embedding space as social media (Reddit, Twitter) comments. 

\textbf{Few shot emotion classification:} Guibon et.al \cite{guibon2021few} explore the problem of emotion classification from conversations in a few shot setting by computing Euclidean distances between the example conversational utterances and emotion-centric prototypes, formed from the average of the word embeddings of utterances. This is the basis for our projection methods, where we project the example and emotion into the same vector space before computing the similarity score. Specifically in our labeled sentences projection method, for each emotion label, we consider embeddings of few labeled examples from the dataset to compute label specific embeddings via averaging.

\textbf{Knowledge transfer for emotion classification:} For multi-turn conversations, Hazarika et.al \cite{DBLP:journals/corr/abs-1910-04980} explore transfer learning from a generative source model trained for dialogue modeling to the target domain of emotion classification from low and moderately sized datasets.
For social media comments and personal reports, Demszky et.al \cite{demszky2020goemotions} perform knowledge transfer from GoEmotions-specific models by finetuning output layers for target datasets through a set of freezing and unfreezing operations. We consider this as the motivation behind using Go-Emotions specific models in our unsupervised methods on Semeval tweets dataset, without going through the additional step of finetuning.


\section{Data}
\label{sec:data}

Tweets and Reddit posts contain expressions of personal opinions and emotions and serve as good test sets for emotion recognition from text.
We chose the GoEmotions Reddit dataset \cite{demszky2020goemotions} and the benchmark SemEval 2018 tweets dataset \cite{mohammad-etal-2018-semeval} for our work.

\subsection{GoEmotions Data}
\label{sec:goemotions}

\begin{figure}[h!]
\centerline{\includegraphics[width=0.45\textwidth]{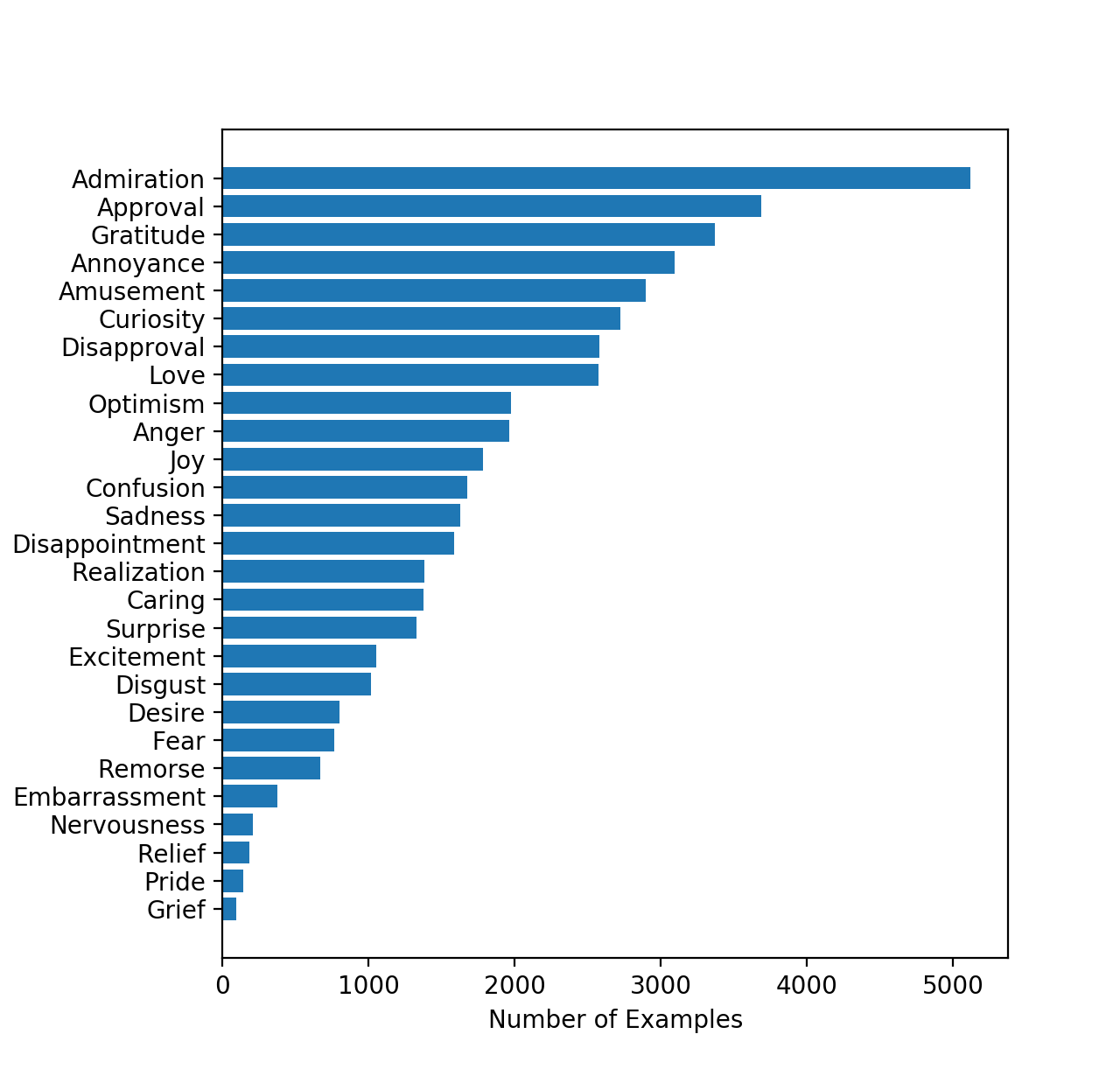}}
\caption{GoEmotions Reddit Emotion Distribution}
\label{Goemotions Dataset distribution}
\end{figure}

The GoEmotions dataset contains 58K Reddit comments, manually annotated for 27 emotion categories.
Each comment can have zero or more emotions labels.
Figure~\ref{Goemotions Dataset distribution} shows the class distribution in this data.
As shown in the figure, the dataset is highly imbalanced, with the largest class (\textit{admiration}) containing 5512 samples versus the 96 samples for the smallest class (\textit{grief}).
The Reddit comments come from the reddit-data-tools project \footnote{https://github.com/dewarim/reddit-data-tools} from the years 2005-2019. 
Subreddits with high levels of profanity or low emotion content were excluded.
The comments are 3 to 30 tokens long, with a median length of 12 tokens. 
At least three raters had annotated each example, achieving significant interrater correlation for each emotion.

\subsection{SemEval 2018 Data}
\label{sec:semeval}

\begin{figure}[htbp]
\centerline{\includegraphics[width=0.45\textwidth]{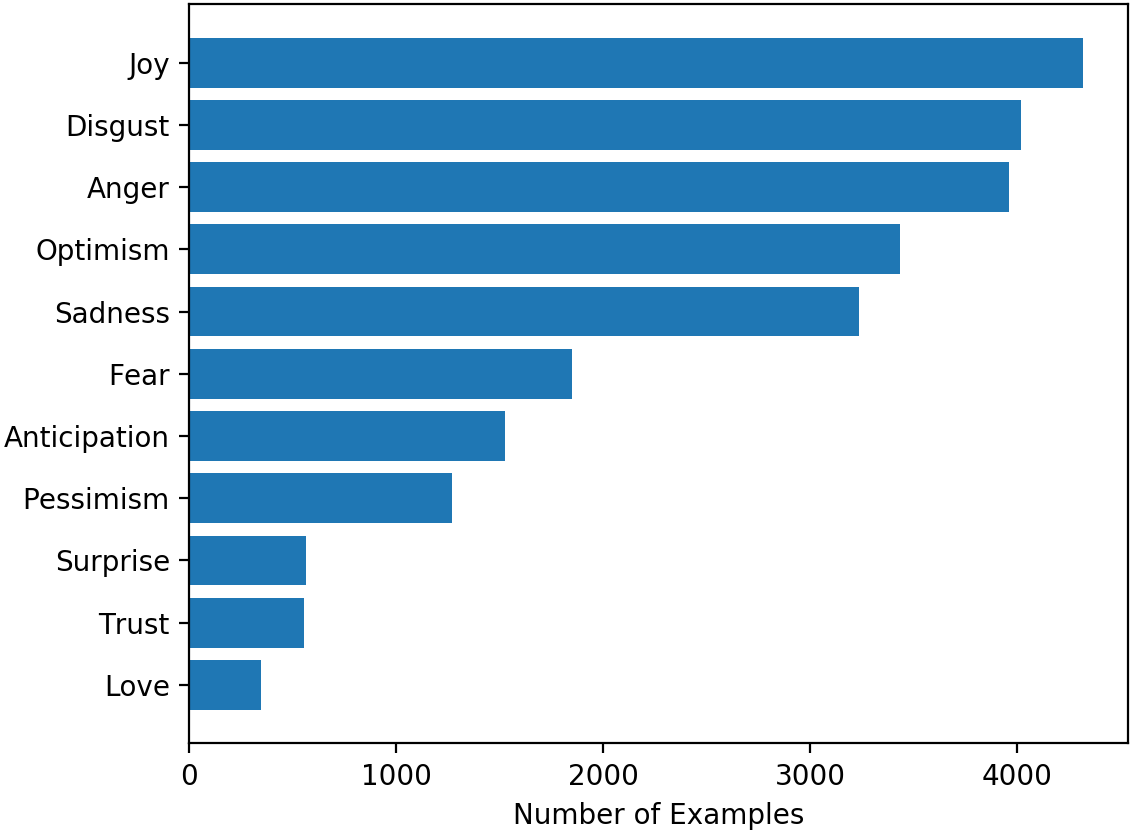}}
\caption{SemEval 2018 Tweets Emotion Distribution}
\label{Semeval dataset}
\end{figure}

We used the English subset of the Affect in Tweets dataset of the 2018 International Workshop on Semantic Evaluation. 
The dataset contains 10.9K Tweets from 2016-2017, annotated for 11 emotion categories. 
Each tweet can have zero or more emotion labels.
Figure \ref{Semeval dataset} shows the class distribution for this dataset.
At least 7 raters annotated each Tweet,  achieving significant interrater correlation for each emotion.
Three emotion labels: \textit{anticipation}, \textit{trust}, and \textit{pessimism}, are absent from the GoEmotions taxonomy.
The comments are 1 to 36 tokens long, with a median length of 16 tokens.
In our experiments, we attempt to transfer knowledge from the GoEmotions Reddit comments to SemEval tweets using unsupervised models.

\section{Methods}

We consider both supervised and unsupervised approaches to model the sentence-level emotion.

\subsection{Supervised methods}
Herein we focus on well-established linear models as well as the more recently proposed transformer models for supervised emotion classification. 

\vspace{0.1cm}
\noindent \textbf{Linear models}:
As simple baselines, we trained logistic regression, linear support vector machines (SVM), and XGBoost classifiers \cite{Chen:2016:XST:2939672.2939785}. 
We used bag-of-ngrams ($n = 2$), LIWC \cite{Pennebaker} category scores, and curated a set of common emoticons and emojis, to construct our feature set. 
We removed punctuations and special characters, lemmatized the word tokens, and excluded ngrams occurring with frequency less than three.
In order to address the class imbalance of the emotion datasets, we weigh the classes according to: $W_e = \frac{N}{(|E|\cdot N_e)}$, where $N$ is the total number of samples, $|E|$ is the number of emotions, and $N_e$ is the number of examples labeled with emotion $e$, using the scikit-learn framework \footnote{https://scikit-learn.org/stable/}.

\vspace{0.1cm}
\noindent \textbf{Transformers}:
We fine-tuned a BERT \cite{BERT} and SBERT \cite{reimers-2019-sentence-bert} based encoder with a dense output layer. 
We used a pretrained BERT-Base model from the huggingface \footnote{https://huggingface.co/transformers/} library.
For SBERT, we used an all-mpnet-base-v2 layer, pretrained using a billion sentence pairs from multiple datasets. 
Both transformer encoders have a hidden state dimension of 768.
We did not perform any text preprocessing. 
In our transformer-based experiments, we calculated the class weights using: $W_{e} = \sqrt{neg_e/pos_e}$, where $neg_e$ and $pos_e$ are the number of negative and positive examples of emotion class $e$, respectively. 
The square root dampens the impact of the negative to positive ratio, which otherwise over-corrects the class imbalance.  
\subsection{Unsupervised methods}
\label{sec:unsupervised}

We propose a method of low resource few shot transfer learning by utilizing the embeddings produced by the transformer models fine-tuned on GoEmotions for an unsupervised classification task on the SemEval dataset. 
The GoEmotions and SemEval datasets have different taxonomies with overlapping labels in the emotion space. 
We leverage the fact that emotion classes not included as a part of training data are words and can be represented in the same word/sentence embedding space as comments. 
We project the sentences (Reddit comments or tweets) and the emotion labels onto the same embedding space.
This allows us to compute the similarity between a given sentence and different emotion classes.
We assign an emotion to a sentence if their cosine similarity score is higher than the emotion-specific threshold.
We use the development set of the target dataset to set the threshold values.
We discuss different projection approaches in the following subsections.

\begin{figure}
\centering
\begin{subfigure}[b]{0.45\textwidth}
   \includegraphics[width=\columnwidth]{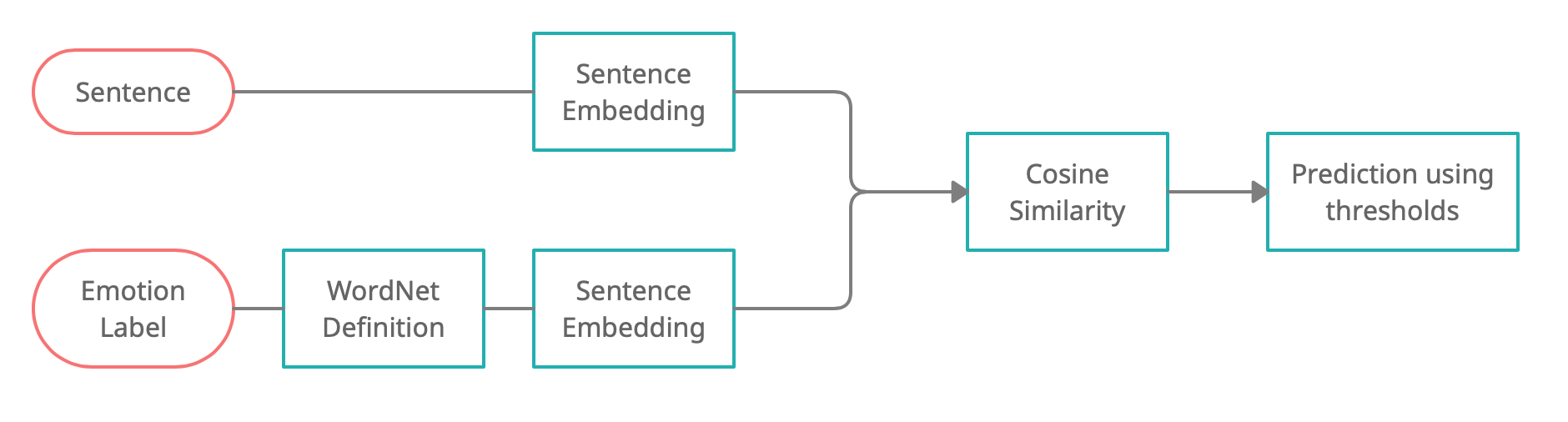}
   \caption{WordNet Definition}
   \label{Sentsim} 
\end{subfigure}

\begin{subfigure}[b]{0.45\textwidth}
   \includegraphics[width=1\columnwidth]{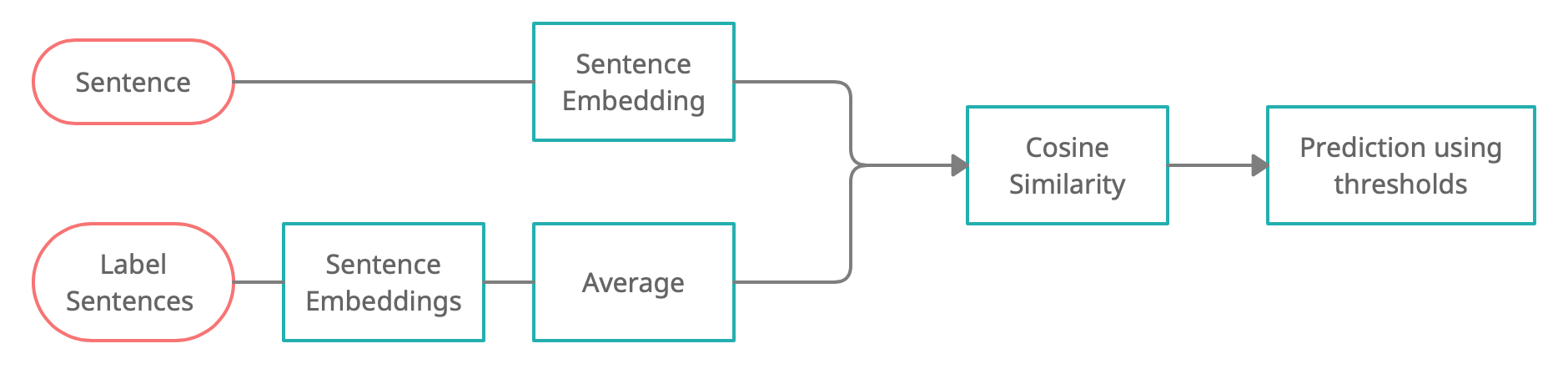}
   \caption{Labeled Sentences}
   \label{Centsim}
\end{subfigure}

\begin{subfigure}[b]{0.45\textwidth}
   \includegraphics[width=1\columnwidth]{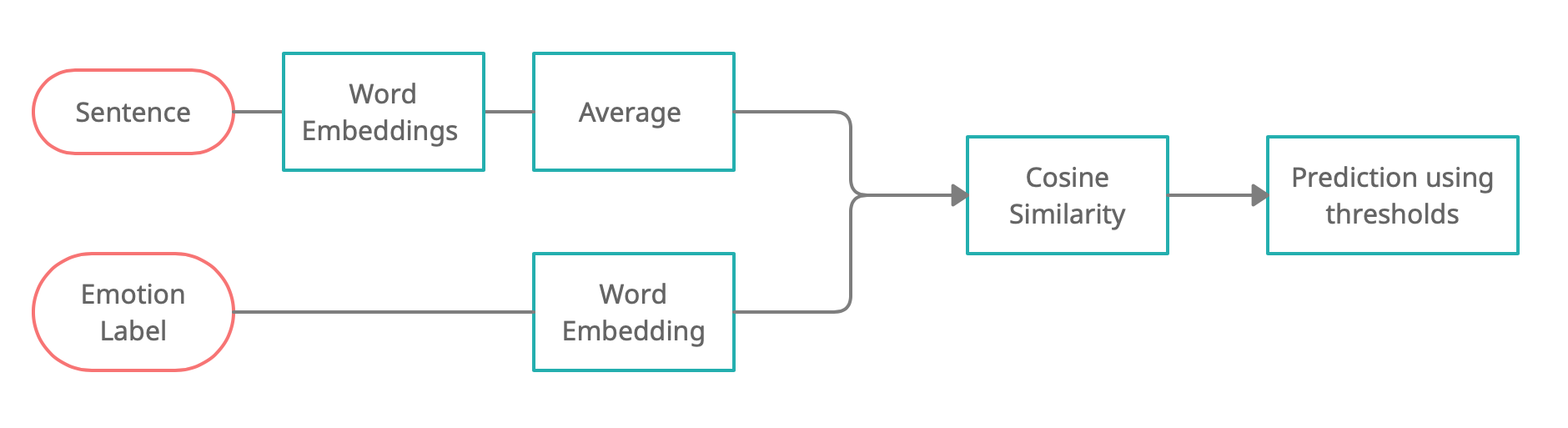}
   \caption{Word Embeddings}
   \label{Wordsim}
\end{subfigure}

\caption[Similarity measures]{Diagrammatic layout of various projection approaches for unsupervised classification}

\end{figure}

\vspace{0.1cm}
\noindent \textbf{WordNet Definition}:
WordNet is a lexical database which arranges words in synonym sets called synsets.
The synsets provide a short sentence defining the meaning of its constituent words.
The synset definition of emotions should be semantically similar to sentences where that emotion is expressed.
Therefore, we find the emotion representation by appending the WordNet definition of the emotion to its lexical form.
For example, we represent the emotion \textit{embarrassment} as ``embarrassment: the shame you feel when your inadequacy or guilt is made public.''
We find the sentence embedding of this emotion definition and the sentence and compute the cosine similarity between them.
Fig.~\ref{Sentsim} illustrates this approach.

\vspace{0.1cm}
\noindent \textbf{Labeled Sentences}:
We find the emotion representation by averaging the sentence embeddings of the development set sentences which are labeled with the corresponding emotion.
We use this to compute the cosine similarity against the sentence embedding of the input sentences, as shown in Fig.~\ref{Centsim}.

\vspace{0.1cm}
\noindent \textbf{Word Embeddings}:
The emotion representation is the word embedding of the emotion label. 
For each input sentence, we remove punctuation and stop words, and average the word embeddings of each token to find the sentence representation, as shown in Fig.~\ref{Wordsim}.

\section{Experiments}

We perform three sets of emotion classification experiments: 1) supervised and 2) unsupervised modeling of GoEmotions Reddit comments, and 3) knowledge transfer from GoEmotions to SemEval using unsupervised approaches.
We use macro precision, recall, and F1 scores to evaluate our models.
We follow the same 8:1:1 train-test-dev split, as used in \cite{demszky2020goemotions}, for the supervised experiments on GoEmotions. 
For the unsupervised and knowledge transfer experiments, we chose smaller development sets than the original splits to emulate few-shot learning.
We used 938 Reddit comments (80\% smaller) and 572 SemEval tweets (35\% smaller) as our GoEmotions and SemEval development sets.

We trained the transformer models using a sigmoid binary cross entropy loss function for 5 epochs and a learning rate of 5e-5.
We used a batch size of 16.
We experimented with two variations of the sentence encoder for the \textbf{WordNet Definition} and \textbf{Labeled Sentences} approach: 1) a pretrained SBERT, 2) SBERT fine-tuned on the supervised emotion classification task on GoEmotions training set.
The latter constitutes a knowledge transfer of emotion-aware sentence embeddings from GoEmotions to SemEval.
We also tried BERT encoders, but SBERT performed better, and we only present its results.
We used 200-dimensional GloVe \cite{pennington2014glove} vectors for the \textbf{Word Embeddings} unsupervised approach.

\vspace{0.1cm}
\noindent \textbf{Baselines}:
We use the BERT-based model from Demszky et al. \cite{demszky2020goemotions} as the GoEmotions baseline.
For the SemEval baseline, we use the winning entry of the corresponding task: NTUA-SLP \cite{NTUA-SLP}.
It used a Bi-LSTM framework with a multi-layer self-attention mechanism.

\section{Results and Discussion}

\subsection{Supervised approaches}
Table \ref{tab:supervised} shows the results of supervised emotion classification on GoEmotions dataset. 
BERT and SBERT achieves similar performance of 0.49 and 0.48 F1 respectively, performing better than the baseline model.
However, their recall is lower than the baseline.
The linear models achieve comparable recall but suffer in precision.
Gradient boosting attained the best precision score of 0.47.

\begin{table}[htbp]
\centering\begin{tabular}{l|ccc}
    Model & Precision & Recall & F1 \\
    \hline
    Baseline \cite{demszky2020goemotions} & 0.40 & \textbf{0.63} & 0.46 \\
    Logistic Regression & 0.30 & 0.57& 0.39 \\
    Linear SVC & 0.29 & 0.61 &0.39 \\
    XGBoost & \textbf{0.47} & 0.46 & 0.44 \\
    BERT & 0.43 & 0.59 & \textbf{0.49} \\
    SBERT & 0.45& 0.57 & 0.48 \\
    \hline
\end{tabular}
\caption{Supervised emotion classification performance on GoEmotions}

\label{tab:supervised}
\end{table}
\vspace{-7mm}
\subsection{Unsupervised approaches}

Table \ref{tab:unsupervised} shows the results of unsupervised emotion classification on GoEmotions dataset.
We use the GoEmotions development set to tune the threshold values, as described in Sec.~\ref{sec:unsupervised}.
The performance of the unsupervised approach is not comparable to the supervised models of Table \ref{tab:supervised}, with the best method (Labeled Sentences) only achieving 0.15 F1.
Replacing the pretrained SBERT encoder with the fine-tuned SBERT layer of the supervised models dramatically improves the performance.
We achieve a maximum F1 score of 0.4 by using this hybrid approach.
The Labeled Sentences approach performs slightly better than the WordNet Definition approach, followed by the Word Embeddings method.
The model is no longer unsupervised because we use the training set to fine-tune the SBERT embeddings.
However, we are not constrained to the dataset's label set and can make predictions for new emotions.
Our experiments on knowledge transfer, described next, supports this.

\subsection{Knowledge Transfer}

Table \ref{tab:knowlege} shows the results of emotion classification on SemEval dataset.
We use the SemEval development set to tune the thresholds for the unsupervised approaches.
The precision and F1 scores increase when we use the fine-tuned SBERT encoder of the GoEmotions-trained supervised models.
Therefore, the knowledge gained by the SBERT embeddings from supervision on GoEmotions improves emotion classification on the SemEval dataset.
We attain a maximum F1 score of 0.49 using the WordNet Definition approach.
This is comparable to the NTUA-SLP supervised baseline, which achieves 0.53 F1.
The WordNet Definition method performs slightly better than the Labeled Sentences approach.

\begin{table}[htbp]
\centering\begin{tabular}{l|ccc}
    Model & Precision & Recall & F1 \\
    \hline
    Unsupervised \\
    \hspace{0.2cm} WordNet + SBERT & 0.11 & 0.33& 0.12 \\
    \hspace{0.2cm} Lbl St  + SBERT & 0.23 & 0.14 & 0.15 \\
    \hspace{0.2cm} Word Em + GloVe & 0.11 & 0.41 & 0.11 \\
    Supervised \\
    \hspace{0.2cm} WordNet + SBERT FT & 0.35 & \textbf{0.51} & 0.38 \\
    \hspace{0.2cm} Lbl St  + SBERT FT & \textbf{0.38} & 0.48 & \textbf{0.40} \\
    \hspace{0.2cm} Baseline \cite{demszky2020goemotions} & 0.40 & 0.63 & 0.46 \\
    \hspace{0.2cm} SBERT & 0.45 & 0.57 & 0.48 \\
    \hline
\end{tabular}
\caption{
Unsupervised emotion classification performance on GoEmotions. 
WordNet = WordNet Definition, Lbl St = Labeled Sentences, Word Em = Word Embeddings (see Sec \ref{sec:unsupervised}).
SBERT = Pretrained SBERT. SBERT FT = SBERT fine-tuned from supervised training on GoEmotions.
}
\label{tab:unsupervised}
\end{table}

\begin{table}[h!]
\centering\begin{tabular}{l|ccc}
    Model & Precision & Recall & F1\\
    \hline
    Unsupervised \\
    \hspace{0.2cm} WordNet + SBERT & 0.30 & 0.60 & 0.39 \\
    \hspace{0.2cm} Lbl St + SBERT & 0.30 & 0.51 & 0.37 \\
    \hspace{0.2cm} Word Em + GloVe & 0.28 & 0.70 & 0.36 \\
    Knowledge Transfer \\
    \hspace{0.2cm} WordNet + SBERT FT & \textbf{0.48} & 0.56 & \textbf{0.49} \\
    \hspace{0.2cm} Lbl St + SBERT FT & 0.40 & \textbf{0.66} & 0.47 \\
    Supervised Baselines \\
    \hspace{0.2cm} Random & - & - & 0.29\\
    \hspace{0.2cm} SVM-Unigrams & - & - & 0.44\\
    \hspace{0.2cm} NTUA-SLP \cite{NTUA-SLP} & - & - & 0.53 \\
 \hline
\end{tabular}
\caption{
Knowledge transfer of emotion classification from GoEmotions to SemEval.
See caption of table \ref{tab:unsupervised} for full form of abbreviations.}
\label{tab:knowlege}
\end{table}

\begin{figure}[h!]
\centerline{\includegraphics[width=0.45\textwidth]{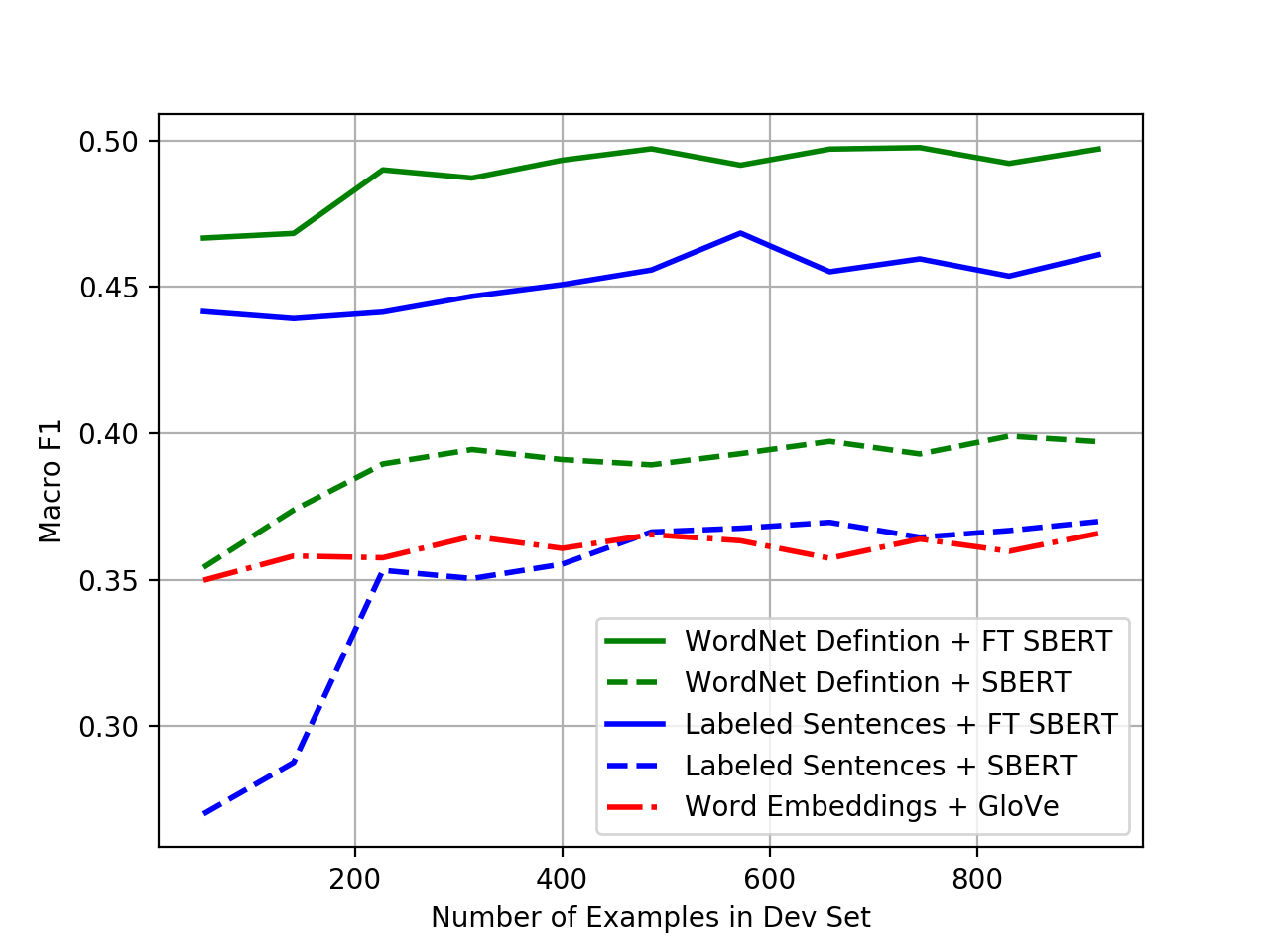}}
\caption{Variation of unsupervised results on SemEval with number of dev set examples. FT SBERT refers to fine-tuned from supervised training on GoEmotions, and SBERT refers to the base SBERT embeddings. }
\label{fig:res}
\end{figure}

The unsupervised approach allows us to make predictions on emotions unseen by the fine-tuned encoder.
For example, the WordNet + SBERT FT model scores 0.13, 0.31, and 0.28 F1 on \textit{trust}, \textit{pessimism}, and \textit{anticipation} emotions, which are not found in the GoEmotions label set (see sec \ref{sec:semeval}).
Fig \ref{fig:res} shows the change in F1 score versus the SemEval development set size, which is used to set the threshold for the unsupervised approaches.
The performance steadies after about 600 examples and does not increase significantly after that.
The WordNet Definition and Word Embedding approaches seem more robust to the number of examples used for tuning the threshold, compared to the Labeled Sentences method.

\vspace{-8mm}
\section{Conclusion}
\label{sec:conclusion}

We presented several unsupervised and few-shot methods for emotion classification leveraging lexical knowledge bases, and evaluated them on the GoEmotions Reddit dataset.
The sentence embeddings fine-tuned on the GoEmotions Reddit comments improved classification performance on the SemEval tweets, with minimal in-domain supervision.
This showed that we could successfully transfer the knowledge gained from supervision on one emotion dataset to another, even if they followed different label taxonomies.
Future work includes using synonym and semantic relations with knowledge graphs to enrich the emotion representations beyond simple definitions.


\end{document}